\documentclass[letterpaper, 10 pt, conference]{ieeeconf}  

\IEEEoverridecommandlockouts                              

\usepackage{graphicx}
\usepackage{subcaption}

\title{\LARGE \bf
UCP-Net: Unstructured Contour Points for Instance Segmentation
}

\author{Camille Dupont$^{*}$, Yanis Ouakrim$^{*}$ and Quoc Cuong Pham \\
\small Université Paris-Saclay, CEA, List, F-91120, Palaiseau, France. Email: \{camille.dupont, quoc-cuong.pham\}@cea.fr\\
\small *Equal contribution
}

\begin{document}


\twocolumn[{%
\renewcommand\twocolumn[1][]{#1}%
\maketitle
\vspace*{-9pt}
\begin{center}
    \centering
    \includegraphics[width=\linewidth]{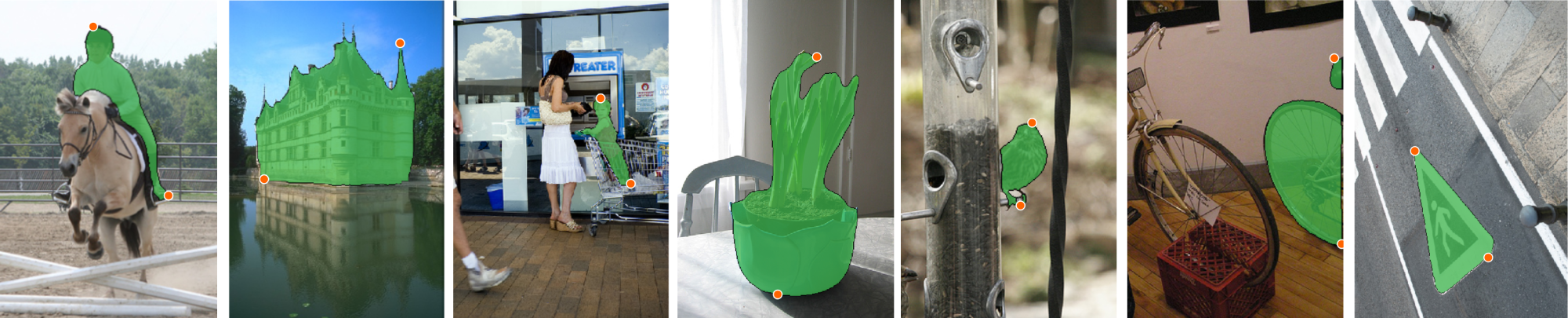}
    \captionof{figure}{Segmentation results of UCP-Net with only two user clicks on the object contour. The model is trained on SBD \cite{bharath11} but is able to perform well for unseen classes such as ground-markings (right).}
    \label{fig:overview}
\end{center}%
}]


\thispagestyle{empty}
\pagestyle{empty}

\begin{abstract}

The goal of interactive segmentation is to assist users in producing segmentation masks as fast and as accurately as possible. Interactions have to be simple and intuitive and the number of interactions required to produce a satisfactory segmentation mask should be as low as possible. In this paper, we propose a novel approach to interactive segmentation based on unconstrained contour clicks for initial segmentation and segmentation refinement. Our method is class-agnostic and produces accurate segmentation masks (IoU $>$ 85\%) for a lower number of user interactions than state-of-the-art methods on popular segmentation datasets (COCO MVal, SBD and Berkeley).

\end{abstract}

\section{Introduction}

\begin{figure*}[!t]
    \centering
    \includegraphics[width=\linewidth]{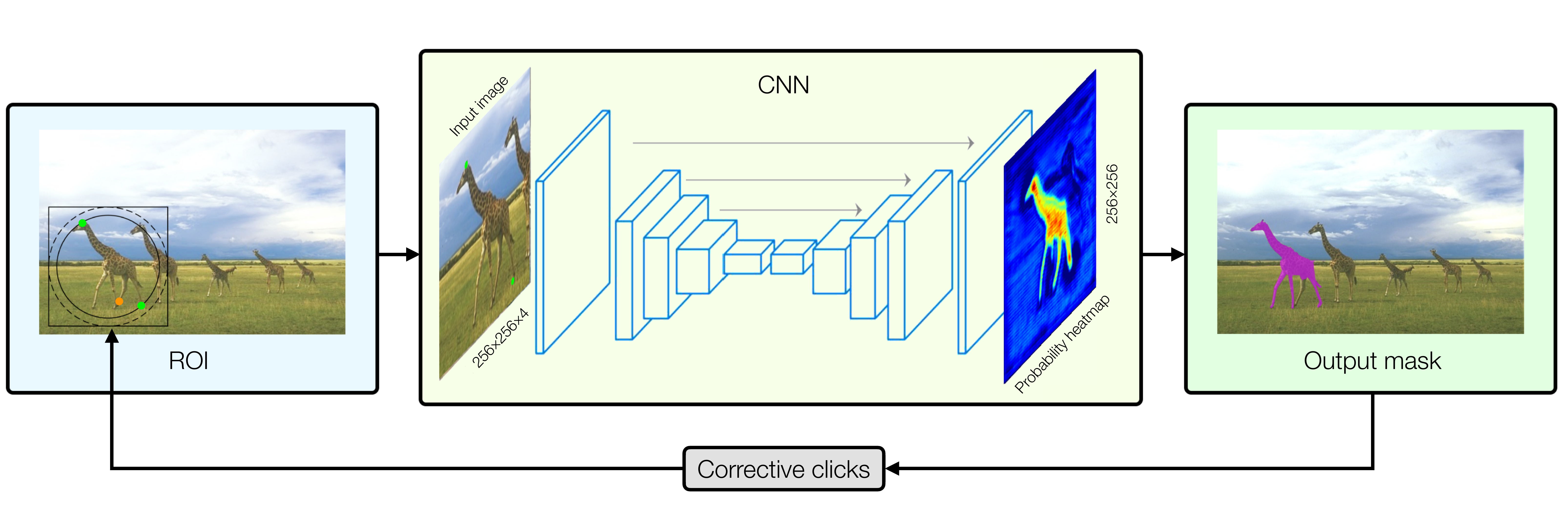}
    \caption{Method overview. Unconstrained user clicks (green dots) are used to crop a Region of Interest, calculated to ensure the object enclosure by solving the smallest-circle problem on SBD train set \cite{bharath11}. Like interactive segmentation methods relying on extreme clicks, the crop is then fed as a fourth image channel to a CNN, illustrated here by U-Net which was used in our experiments. Through a unified approach, UCP-Net supports additional corrective clicks (orange dot) which can be used to refine the predicted segmentation mask.}
    \label{fig:overview}
\end{figure*}
\label{sec:intro}
As deep learning gains popularity, the need for large amounts of annotated images has never been greater.  Annotating images is a tedious and time consuming task, especially in the field of image segmentation where human annotators have to draw complex polygons around all sorts of objects. Interactive image segmentation aims to reduce the workload required to extract objects or regions from images. It relies on sparse user interactions such as clicks or scribbles to produce dense binary masks that precisely encompass the desired regions. It is often an iterative process, where users interact with the algorithm both to initialize and adjust the generated segmentation masks. To be considered effective, an interactive segmentation algorithm must comply with three main requirements: i) meet high-quality standards, i.e. 80-90\% intersection over union (IoU); ii) be less time-consuming than manual segmentation; and iii) be robust to variation in user interactions. It is also usually expected to be robust to domain or category shift.
\bigbreak

Although most approaches rely on positive and negative clicks \cite{xu16, li18, mahadevan18, majumder19}, recent studies have shown that extreme clicks \cite{papadopoulos17,maninis18} can be effectively used to give scale information and indicate precisely points that belong to the object thus removing ambiguities. In particular, the scale information allows the crop of the original image and therefore a higher resolution which significantly increases the performance compared with the full image \cite{maninis18}.
However, such an interaction requires users to click at exact object locations, a time-consuming task which is prone to user inattention mistakes, and to click 4 times regardless of the object complexity, which can either be insufficient at times and dispensable at others.
Our method aims to extend extreme clicking toward generic contour clicking by removing its two main constraints: the fixed number of clicks and the need for specific click location. The key contributions of our work can be summarized as follows:
\begin{itemize}
  \item We design a novel interactive segmentation pipeline suitable for consuming unconstrained contour clicks, hence relaxing the strong cognitive load of extreme points while maintaining the benefits of enclosure-based interactions such as a higher resolution and scale-information.
  \item We show how such a pipeline can be trained to perform satisfactory segmentation from unstructured contour points as few as two.
  \item We conduct an extensive study of main enclosure based interaction types with human annotators.
\end{itemize}

The resulting model is able to perform segmentation in real-time directly in a web browser (53 ms for 128 x 128 instances through WebGL) with no need for a dedicated GPU. We believed that it will constitute a more realistic solution in comparison with the standard deep learning frameworks that require standalone graphics cards.

\section{Related work}

\paragraph{User interaction types} Early approaches to interactive segmentation rely on low-level image features such as pixel color to infer boundaries, thus requiring samples of the foreground and background pixels \cite{greig89, mortensen95}. This often translates into scribble interactions on both foreground (positive) and background (negative) pixels or rough drawing around the target. This information is then fed to a heuristic algorithm to produce a segmentation. The arrival of deep neural networks able to extract higher-level features enabled for sparser interactions such as simple clicks. While most approaches use positive and negative clicks \cite{xu16, li18, mahadevan18, majumder19}, recent studies have shown that extreme clicks \cite{papadopoulos17,maninis18} can be effectively used to give scale information and indicate precisely points that belong to the object thus delivering valuable information. As users must click on the left-most, right-most, top and bottom points of the object they want to segment, the interaction is also more consistent and reliably reproducible. Unlike positive and negative clicks, extreme points have not been extended to an iterative refinement training scheme.

\paragraph{Interaction simulation} In addition to the interaction effectiveness, the mechanism behind the automatic simulation of extreme clicks is a confounding factor for both training and evaluation. Maninis \textit{et al.} \cite{maninis18} observe in the case of extreme clicks a decrease of up to 5\% of the mean IoU between the simulated and real clicks evaluation. We briefly present commonly used simulation strategies to mimic human behavior. Foreground clicks are usually constrained to cover the central area by using a margin from the object boundary or by applying k-medoids \cite{jang19}, whereas negative clicks are either peripheral to the object or on negative objects \cite{xu16, liew17, benard18, li18, liew19, majumder19}.  Stricter interaction policies such as bounding boxes and extreme points simply include noise by perturbing the corners of the perfect coordinates up to a certain pixel amount \cite{maninis18, benenson19, shahin20} or scale percentage \cite{wang19}.

\paragraph{Embedding User Interactions} User interactions being sparse, they require an effective pre-processing so as to be fully perceived and exploited by the segmentation network. A popular pre-processing consists in encoding the interactions into a 2d-image that can be fed to the convolutional network. Clicks are usually turned into Euclidean \cite{xu16,liew17,hu18,li18,jang19} or Gaussian \cite{benard18,li18,maninis18,mahadevan18, forte20} distance maps.
The authors of \cite{benard18, mahadevan18, maninis18, benenson19} observed that Gaussians yield better results than distance transforms. Three other transforms led to an improvement over Gaussians: binary disks \cite{benenson19}, superpixels \cite{majumder19} and multi-focal ellipses \cite{shahin20}, but no comparison between them was provided.

In 2016, Xu et al. \cite{xu16} proposed the first interactive segmentation pipeline relying on a Fully Convolutional Network (FCN) encoder-decoder taking the concatenation between the RGB image and the embedded user interaction as input. Most modern approaches to interactive segmentation follow this lead \cite{mahadevan18, benard18, maninis18, benenson19, majumder19, liew17, liew19, shahin20}. While the majority use the whole image as input \cite{mahadevan18, benard18, majumder19, liew17, liew19, forte20, jang19}, recent architectures based on object enclosure \cite{maninis18, wang19, shahin20, benenson19, zhang20, sofiiuk20} feed image crops to the FCN to achieve speed-up and preserve object details. The approach proposed by \cite{sofiiuk20} takes the whole image as input and exploits the predicted mask boundaries to obtain a crop of the image, which is subsequently fed into a refinement model. Instead of using image patches, Liew \textit{et al.} \cite{liew17} crop the feature maps around the input clicks to infer local predictions which are reassembled afterwards.

In order to learn deep features for images and interaction maps individually, the authors of \cite{hu18, forte20} use two separate encoder streams: one for the image and another one for the interactions, leading however to a heavier model. 

In comparison with negative and positive clicks, extreme clicks have the advantage of being less ambiguous and enable to reduce the search space by extracting an RoI around the object. However, such methods require users to click at exact object locations which is more constraining than positive and negative clicks. Moreover, they require users to click \textit{at least} 4 times regardless of the object complexity. To solve these two main limitations, we propose a novel interactive segmentation approach that exploits unconstrained contour clicks ranging from 2 to $n$. This increased flexibility enables the unified approach to generate masks with different precision levels. We demonstrate that our method is able to deliver high-quality results with a lower number of clicks than the current state of the art of interactive segmentation.

\section{Method}

\begin{figure}
    \centering
    \includegraphics[width=0.8\linewidth]{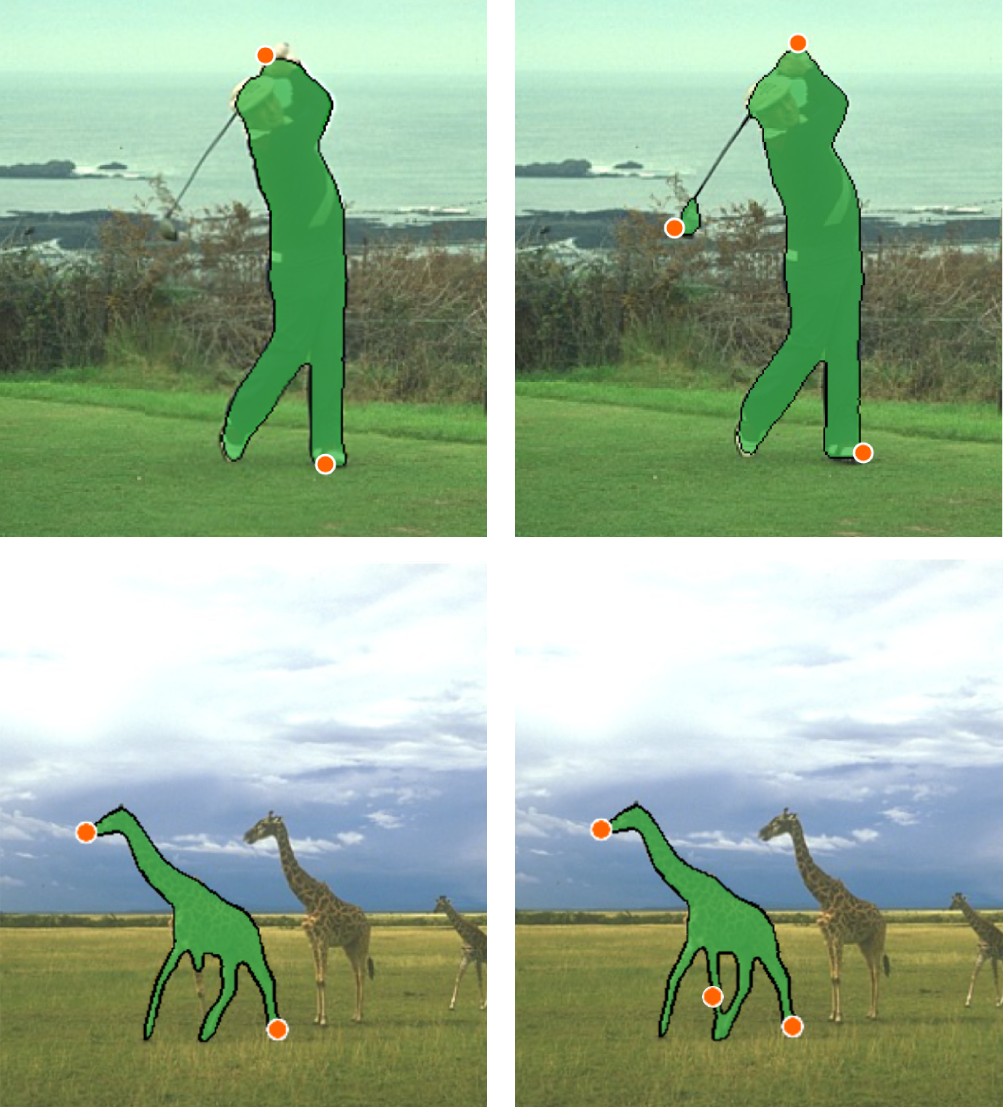}
    \caption{Two example cases where a corrective click is added: target ambiguity (top) and model failure (bottom).}
    \label{fig:corrective_click}
\end{figure}

Our network is built upon both approaches based on extreme clicks and those based on positive negative clicks. Exploiting the contour clicks representing the target object enables us to crop the original image and benefit from a higher resolution. Similar to \cite{maninis18}, the crop is concatenated with its corresponding click heatmap and then fed to a binary segmentation network (Figure \ref{fig:overview}). However, unlike extreme clicks methods and similar to positive-negative methods, we choose to investigate a much broader range of number of clicks with unconstrained locations in an iterative fashion. This flexibility speeds up the interaction process even further and adapts well to both coarse and fine objects. Indeed, we observed that, in most cases two unconstrained user clicks provide enough information for a model to predict an accurate segmentation mask (Table \ref{tab:result_table}). In some cases, complex objects or situations can lead to the two clicks being insufficient for the model to correctly segment the object (Figure \ref{fig:corrective_click}). In regard of this observation, we propose an iterative approach where correction clicks are added until a satisfactory segmentation mask is predicted by the model. Therefore, the number of clicks fits the complexity of the setup, thus speeding up the annotation process.


From a user's perspective, our interactive segmentation pipeline can be summarized as follows: first the user clicks on two locations of the object contour, then they can add additional contour clicks to correct or refine the mask (Figure \ref{fig:overview}).

\paragraph{Simulating user interactions}

\begin{figure}
    \centering
    \begin{subfigure}{.52\linewidth}
        \centering
        \includegraphics[width=\linewidth]{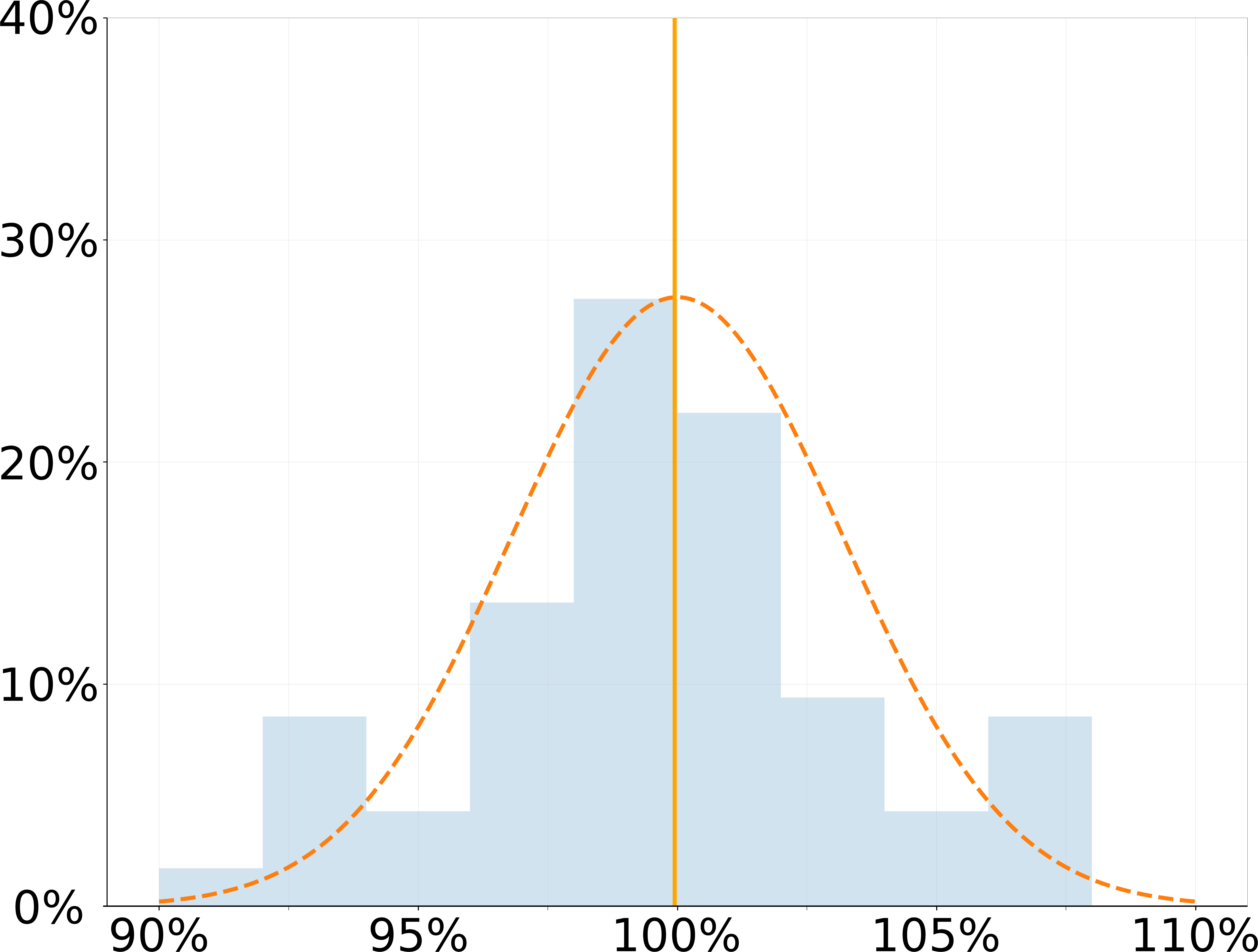}
    \end{subfigure}
    \begin{subfigure}{\linewidth}
        \centering
        \includegraphics[width=\linewidth]{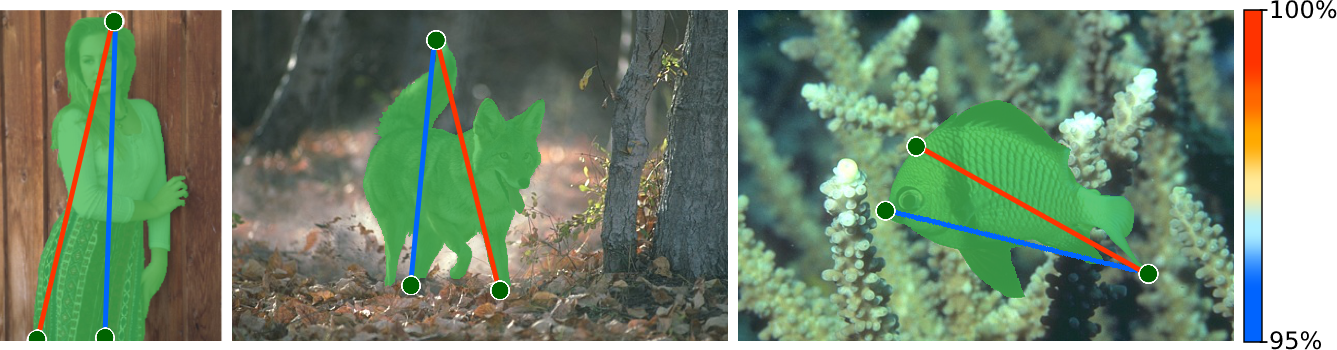}
    \end{subfigure}
     \caption{User pair clicks distance distribution for $N=5$ annotators on the Berkeley dataset \cite{martin01} (top). Visual examples of clicks pairs corresponding to 0.95 and 1 distance ratios with respect to the maximum distance (bottom).
     }
    \label{fig:clicks-distribution}
\end{figure}

Simulating user interactions is a challenge in the field of interactive segmentation. We propose a novel online iterative training scheme, during which our model is trained with a combination of three strategies to simulate human contour clicks. 
When asking 5 annotators to draw a few clicks on object contours on the Berkeley data set (100 instances), we observe that they instinctively distribute them to best represent the targeted object breadth. In particular for $n=2$ contour clicks, we measure that the distance ratio between clicked pair points and the furthest ground truth pair points is approximately distributed as a normal random variable with mean 1 and standard deviation 0.03 (Figure \ref{fig:clicks-distribution}). The interaction can therefore be simulated by selecting pairs producing a distance following this distribution, both during training and testing.\\

\begin{figure}
    \centering
    \begin{subfigure}{.49\linewidth}
        \centering
        \includegraphics[width=.95\linewidth]{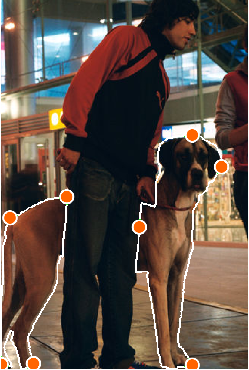}
        \caption{Geometric strategy}
    \end{subfigure}
    \begin{subfigure}{.49\linewidth}
        \centering
        \includegraphics[width=.95\linewidth]{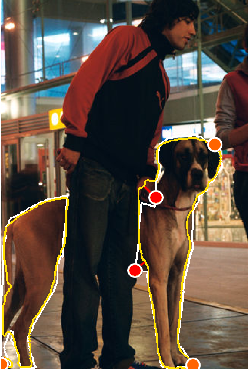}
        \caption{Corrective strategy}
    \end{subfigure}
     \caption{Visual example of two strategies for contour clicks simulation. Geometrical simulation (a) aims to gradually refine salient regions of the target (white line). Corrective strategy (b) simulates an iterative user-like behavior where the user clicks on the contour further (red dots) from the previous prediction (yellow line).
     }
    \label{fig:strategies}
\end{figure}

We describe here the other two simulation strategies for $n > 2$ as illustrated in Figure \ref{fig:strategies}. Let $\mathcal{C}^{gt}$ be the set of ground truth contour pixels.
\begin{itemize}
  \item \textbf{Geometric strategy:} gradually refines salient regions of the target. We denote $\mathcal{C}^{geo}$ the contour pixels resulting from the conversion of the $n-1$ points set into polygon boundaries. The $n^{th}$ click is then obtained sequentially as the furthest ground truth pixel from $\mathcal{C}^{geo}$ so as to mold the clicks to the shape of the target:
  \begin{equation}\label{eq:strat2}
  p^{n}=arg \max_{p \in \mathcal{C}^{gt}} \left( \min_{q \in \mathcal{C}^{geo}} \| p - q \|\right),\ n > 2
  \end{equation}
  \item \textbf{Corrective strategy:} relies upon the prediction of the interactive segmentation network from $n-1$ clicks. We note $\mathcal{C}^{pred}$ its corresponding contour pixels. The $n^{th}$ click is defined as the furthest ground truth pixel from the prediction:
  \begin{equation}\label{eq:strat2}
  p^{n}=arg \max_{p \in \mathcal{C}^{gt}} \left( \min_{q \in \mathcal{C}^{pred}} \| p - q \|\right),\ n > 2
  \end{equation}
  Batch of $\Delta n$ new clicks can be added at once by partitioning the erroneous areas and applying the strategy to each blob.
\end{itemize}

The first aims to best represent the targeted object by gradually refining salient regions. The second aims to simulate human correction to the network errors by selecting contour clicks furthest from the prediction contours. The corrective strategy applies natively to multi-region objects or objects with holes as it is based on euclidean distance from contours regardless of the contours hierarchy. The extension of the geometrical strategy to multi-regions is defined using a coarse-to-fine policy by prioritizing exterior hull coverage and subsequently interior regions as shown in Figure \ref{fig:strategies}.

\paragraph{Region of interest}

\begin{figure}
    \centering
    \includegraphics[width=\linewidth]{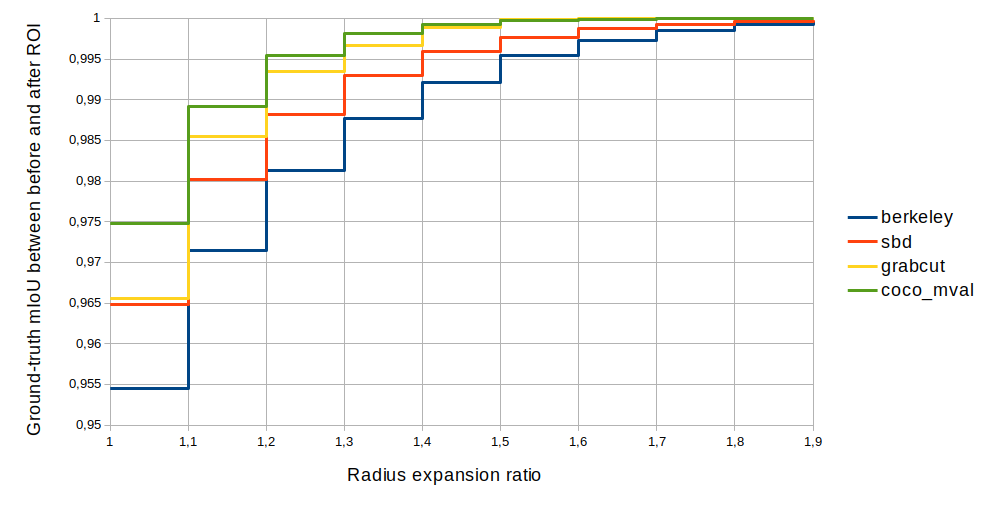}
    \caption{Mean IoU between ground truth mask and itself when cropped using the smallest-circle method enlarged with expansion ratios ranging from 1 to 1.9. The $n=2$ clicks are simulated using max-distance ratio of 0.95. We choose the cut-off value $r_{opt}=1.4$.}
    \label{fig:cutoff}
\end{figure}

Similarly to \cite{mahadevan18}, we feed the network with a crop of the original image to benefit from a higher resolution. While extracting a region of interest is straight-forward in the case where provided clicks give a good approximation of the shape of the targeted object, it can be more difficult in the case of very sparse contour clicks (under four). As described previously, we conducted a human experiment that showed an innate distribution of the clicks to best represent the breadth of the targeted object. To ensure full enclosure of the targeted object, we therefore extract the RoI by solving the smallest-circle problem. To do so, we rely on Welzl’s algorithm \cite{welzl91}. It corresponds to the diagonal's circle and the circumscribed circle for two and three points respectively (Figure \ref{fig:overview}). In order to reckon the users' interaction fluctuation, guarantee object enclosure and have context information, we expand the circle's diagonal by 1.4 (Figure \ref{fig:cutoff}). The crop generated by this cut-off expansion ratio ensures a negligible mean loss ($<$1\%) on more than 20K instances in the SBD train set \cite{bharath11} using simulated pair clicks of 0.95 fraction.

\paragraph{An iterative training scheme}

\begin{figure*}
    \centering
    \includegraphics[width=\linewidth]{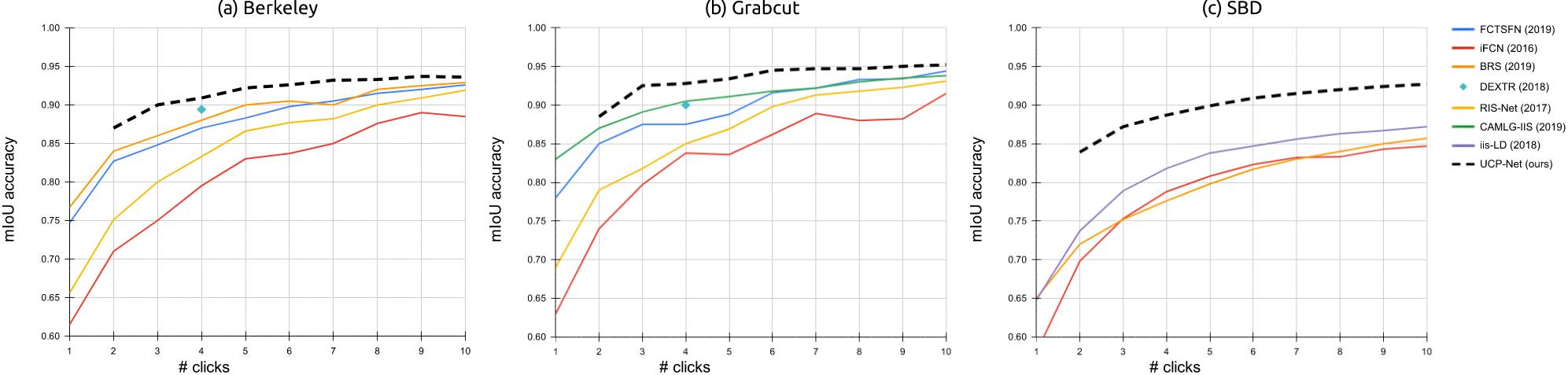}
    \caption{Curves of mean IoU scores after n clicks for Grabcut \cite{rother04} (a), Berkeley \cite{martin01} (b) and SBD \cite{bharath11} (c) test sets. Note that we excluded the out-performing method \cite{forte20} on Grabcut which uses additional synthetic data during training. }
    \label{fig:curves}
\end{figure*}

The training first consists of a warm-up phase with two contour clicks as input. These clicks are simulated geometrically as described in the previous section. During a second stage, we aim to cover a wider range of click numbers and randomly pick $n_{add}$ additional geometric contour clicks to each sample. Experimentally, we observe that a range $n_{add} \in [0,8]$ allows for precise segmentation masks (Figure \ref{fig:curves}).



\begin{table*}
\centering
\begin{tabular*}{\textwidth}{l @{\extracolsep{\fill}} ccccc}
\hline
Method & \begin{tabular}[c]{@{}c@{}}Train\\ Set\end{tabular} & \begin{tabular}[c]{@{}c@{}}SBD\\ @85\%\end{tabular} & \begin{tabular}[c]{@{}c@{}}GrabCut\\ @90\%\end{tabular} & \begin{tabular}[c]{@{}c@{}}Berkeley\\ @90\%\end{tabular} & \begin{tabular}[c]{@{}c@{}}COCO MVal\\ @85\%\end{tabular} \\ \hline \hline
VOS-Wild* \cite{benard18} (2017) & SBD Full & - & 3.8 & - & -  \\
DEXTR* \cite{maninis18} (2018) & SBD Full & - & 4.00 & 4+ (89.4\%) & 4+ (80.1\%)  \\
CAMLGIIS \cite{majumder19} (2019) & SBD Full & - & 3.58 & 5.60 & -  \\
ITIS \cite{mahadevan18} (2018) & SBD Train + VOC12 & - & 5.60 & - & -  \\
IIS-LD \cite{li18} (2018) & SBD Train & 7.41 & 4.79 & - & 7.86 \\
BRS \cite{jang19} (2019) & SBD Train & 6.59 & 3.60 & 5.08 & - \\
f-BRS-101 \cite{sofiiuk20} (2020) & SBD Train & 4.81 & 2.72 & 4.57 & - \\
GAIS \cite{forte20} (2020) & SBD Train + synt. & 3.90 & 2.54 & 3.53 & - \\ \hline
iFCN \cite{xu16} (2016) & VOC12 & 9.22 & 6.08 & 8.65 & 9.07 \\
RIS-Net \cite{liew17} (2017) & VOC12 & - & 5.00 & 6.03 & - \\
FCTSFN \cite{hu18} (2019) & VOC12 & - & 3.76 & 6.49 & 9.62 \\ \hline
MultiSeg** \cite{liew19} (2019) & SBD+VOC & - & \textbf{2.30} & 4.00 & - \\
FAIRS \cite{shahin20} (2020) & VOC12 & 4.0 & 3.0  & 4.0 & - \\ \hline \hline
\textit{\textbf{UCP-Net* (Ours)}} & SBD Train & \textbf{2.73} & 2.76 & \textbf{2.70} & \textbf{2.00} \\ \hline

\end{tabular*}
\caption{Comparison table. mIoU is specified for methods which did not reach the @x\%. *Methods relying on contour clicks. **Using different types of interactions, the authors gave a NoC equivalent of their result.} 
\label{tab:result_table}
\end{table*}

\section{Experiments}

\subsection{Datasets}

We evaluate our model across five publicly available segmentation datasets. To compare our model with other segmentation methods, we use the mean number of clicks necessary to reach the typically used 85-90\% IoU threshold, known as the Number of Clicks metric (NoC @x\%). Forte \textit{et al.} \cite{forte20} argue this widely used metric fails to characterize the ability of models to progress over a wider range of clicks, particularly useful for applications with high-quality requirements such as image editing. They recommend the additional use of accuracy score across a range of clicks.
We use the SBD dataset \cite{bharath11} to train the proposed model. It includes 8,498 training images and 2,857 test images, corresponding respectively to 20,164 and 6,671 instances.

To simulate user clicks during evaluation, we first generate 2 clicks on the target object and then apply the corrective strategy to refine the prediction. To compute the NoC@x\% metric, the refinement is limited to the targeted IoU threshold. To compute the mIoU for progressive $n$ clicks, the refinement is limited to $n$ clicks.

\subsection{Comparison of user interactions}
 To compare contour clicks with traditional enclosure interactions, we conducted an experiment with five human annotators. Annotators had to label the 100 images of the Berkeley dataset \cite{martin01} using bounding boxes, extreme clicks, as well as three free contour clicks and two free contour clicks.
\begin{table}
    \centering
    \begin{tabular}{lccc}
        \hline
       Interaction type &
       NoC &
       Average time (s) &
       Median time (s) \\
        \hline
        \hline
        Extreme clicks & 4 & 9.13 & 8.51 \\
        Bounding boxes & 2 & 6.37 & 6.18 \\
        Free contour clicks & 3 & 5.51 & 5.42 \\
        Free contour clicks & 2 & 3.78 & 4.36 \\
        \hline
    \end{tabular}
    \caption{Interaction time for extreme clicks, bounding boxes and unconstrained contours clicks on the 100 Berkeley images \cite{martin01} ($N=5$ annotators).}
    \label{tab:time_comparison}
\end{table}

\begin{table}
    \centering
    \begin{tabular}{lccc}
        \hline
       Interaction type &
       NoC &
       Simulated clicks &
       Real user clicks \\
        \hline
        \hline
        Extreme clicks \cite{maninis18}* & 4 & 89.1 & 87.7 \\
        Free contour clicks & 2 & 87.0 & 86.3 \\
        \hline
    \end{tabular}
    \caption{mIoU comparison between extreme clicks (DEXTR \cite{maninis18}) and unconstrained contours clicks using either simulated clicks or real clicks on the 100 Berkeley images \cite{martin01}. * Re-implemented using a EfficientNet-B6 backbone, so that only the user interaction varies.}
    \label{tab:extreme_comparison}
\end{table}

\begin{figure}
    \centering
    \includegraphics[width=\linewidth]{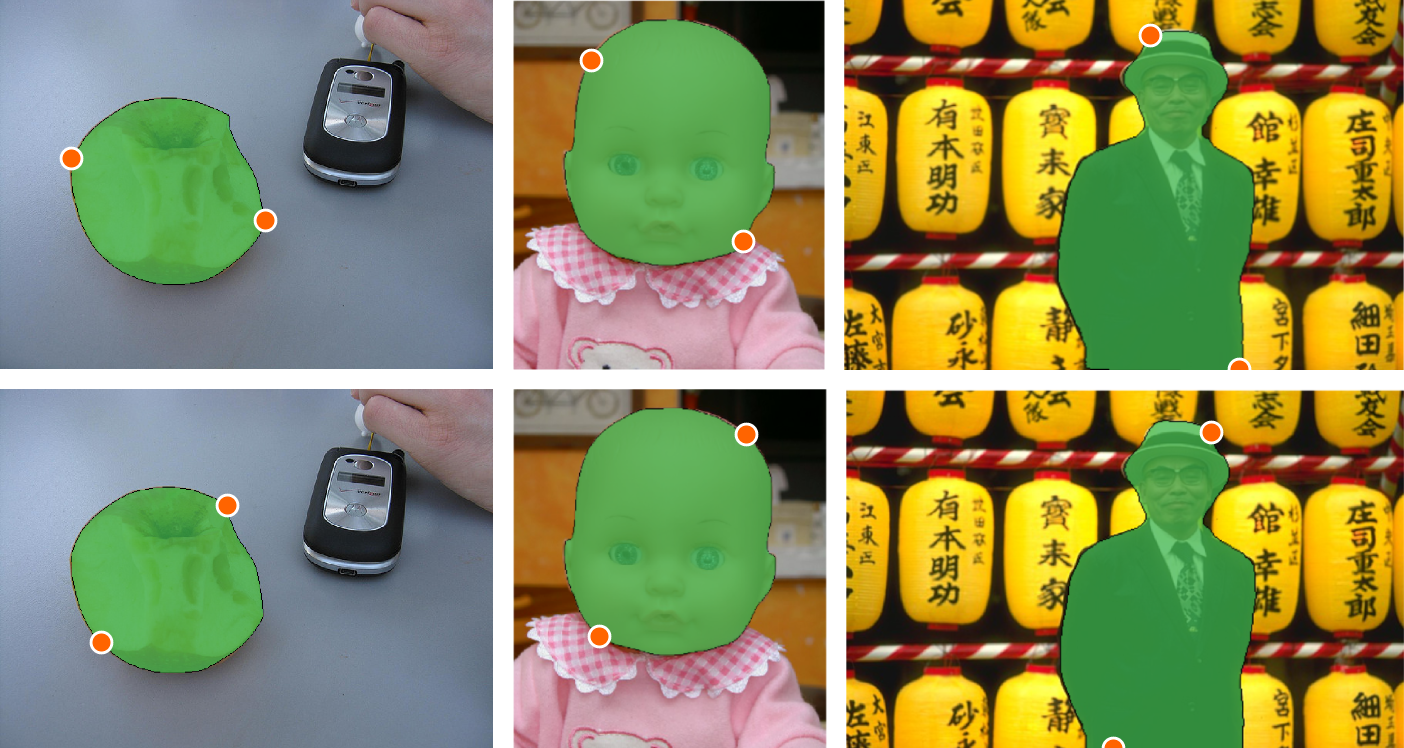}
    \caption{Location noise during training allows for a robustness to user variation.}
    \label{fig:noise}
\end{figure}

\begin{figure*}
    \centering
    \includegraphics[width=\linewidth]{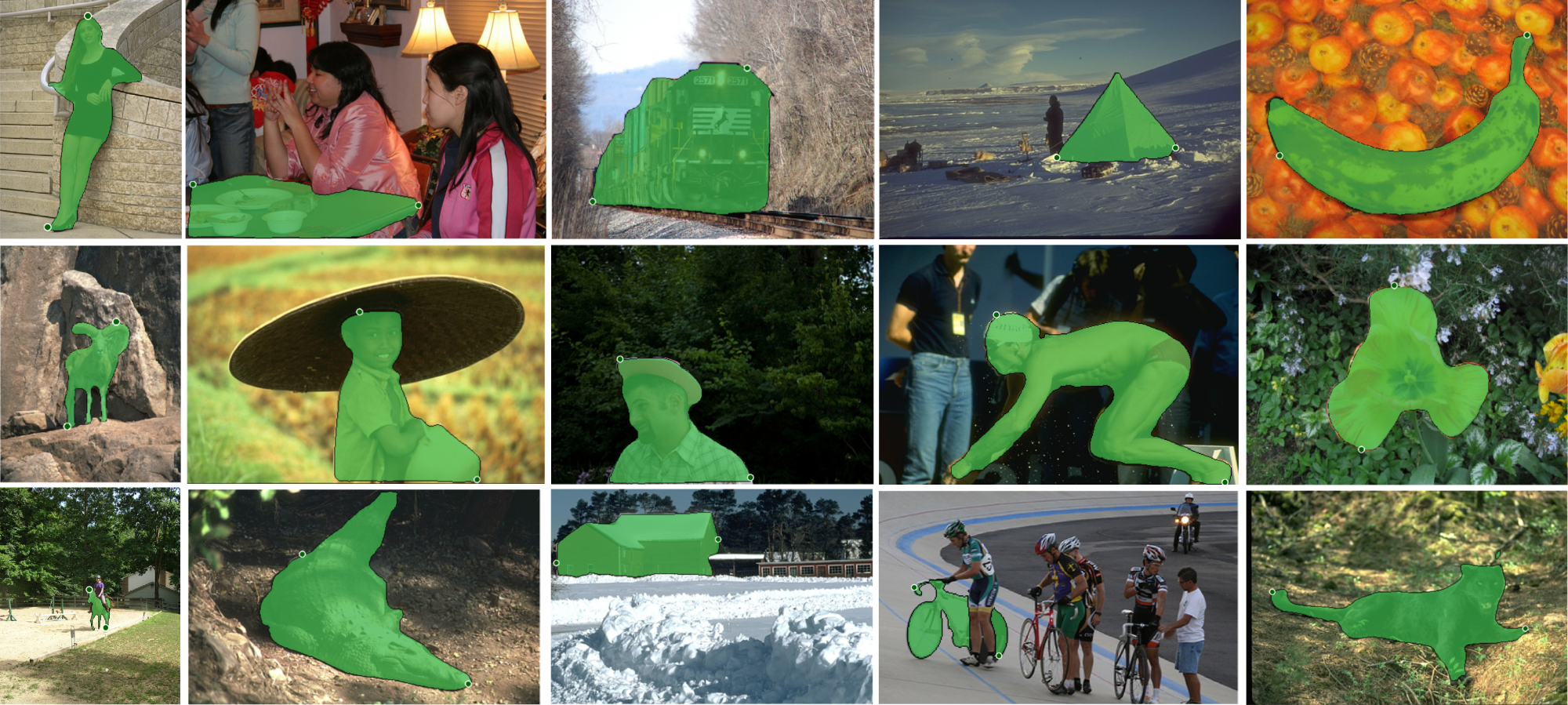}
    \caption{Examples of predictions on the Berkeley dataset \cite{martin01} with their corresponding input clicks.}
    \label{fig:predictions}
\end{figure*}
\begin{figure}
    \centering
    \includegraphics[width=0.9\linewidth]{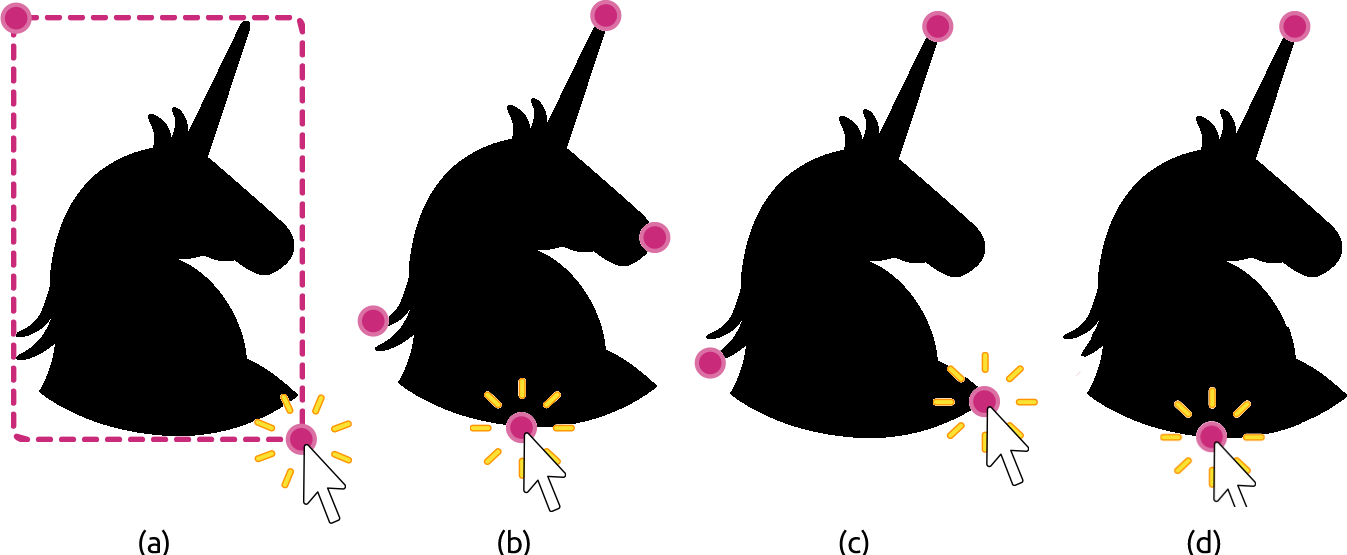}
    \caption{Annotation visual guidelines given during the time comparison experiment for (a) bounding box, (b) extreme points, (c,d) free contour clicks.}
    \label{fig:illustration_guidelines}
\end{figure}
\begin{figure}
    \centering
    \includegraphics[width=\linewidth]{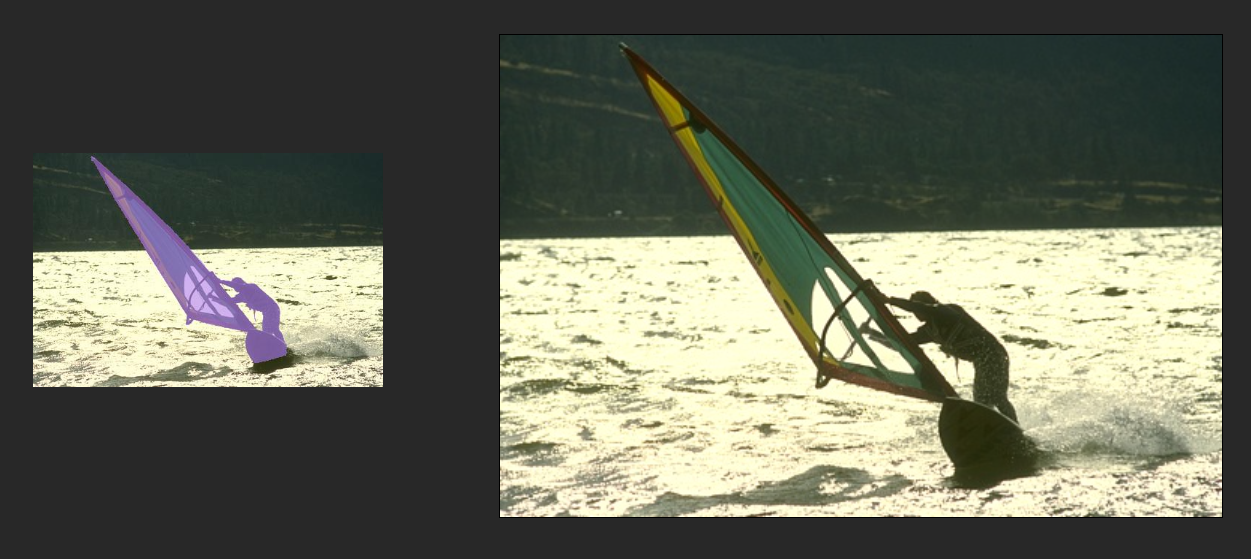}
    \caption{Experiment interface. The target object is shown briefly during 2s (left), with a reduced image size to prevent the user from anticipating the cursor position. Afterwards, the image is displayed at full resolution (right).}
    \label{fig:interface}
\end{figure}

The user guidelines for each interaction types are shown in Figure \ref{fig:illustration_guidelines}. To both designate the object of interest in the image and prevent the users from anticipating their cursor position, a miniature of the image with the ground truth mask of the targeted object is briefly displayed during two seconds. Then the miniature is replaced with the full resolution image on which the user can draw a box or click on contour points. Figure \ref{fig:interface} gives an overview of the annotation interface. While click precision was not mentioned in the guidelines, we calculated the accuracy with respect to the ground-truth box to ensure fairness in time comparison between extreme points and bounding boxes and found no significant difference in standard deviation (3.2\% vs 3.6\%).

Results are shown in Table \ref{tab:time_comparison}. Two user-clicks proved to be almost three times faster than extreme clicks, while also being significantly faster than simple bounding boxes. Note that this finding is contradictory with the results of Papadopoulos \textit{et al.} \cite{papadopoulos17} who observed that extreme points (7.2s) were significantly faster than bounding boxes (34.5s).

\subsection{Implementation details}

\paragraph{Architecture and hyper-parameters} 
Like many previous interactive segmentation methods \cite{bredell18}, we use a U-Net \cite{ronneberger15} architecture. We replace the VGG backbone with an EfficientNet-B6 \cite{tan19} which has become the backbone of choice for many deep learning tasks and is at the top of the ImageNet classification leader-boards. After a pre-training on ImageNet, we train on SBD train (20,172 instance images; 8,498 images) and use the SBD val for validation (6,671 instance images). Simulated user clicks are represented as Gaussian distance functions and fed to the model as a fourth image channel. Unless specified otherwise, the results given in this report were obtained while evaluating on SBD val.
We use the dice coefficient as our loss function as our experiments demonstrated that it enables a slightly higher mean IoU than binary cross entropy alone and binary cross entropy and the dice coefficient combined. We use a learning rate of 1e-5, that is reduced to 1e-6 when the loss has not been improving for the last 15 epochs. Training stops after 7 epochs without improvement of the loss function. We use a batch size of 12 as it gave better results than batch sizes of 8 and 16. We resize images to 256*256 as it enables to obtain a better IoU than resizing the images to 128*128 or 512*512. We set dropout to 0.5.

\paragraph{Data augmentation} As genericity is a critical component in annotation assistance systems, we apply a substantial variety of image augmentations to our images during the training phase. We rely on the ImgAug \cite{imgaug} library to apply noise (drop out, coarse drop out, Gaussian noise, weather changes...), color changes (gama contrast, temperature...) and geometric transformations (perspective, vertical and horizontal flips...) with default parameters. In terms of user interaction, we add Gaussian noise to the simulated clicks as shown in Figure \ref{fig:clicks-distribution}.

\subsection{Evaluation}

Table \ref{tab:result_table} provides a comparison of UCP-Net against previous interactive segmentation methods. We reach standard benchmark IoUs with lower numbers of clicks on SBD and Berkeley while getting close to state of the art on GrabCut. Moreover, we conducted an ablation study to evaluate the accuracy gain between extreme and unconstrained contour points. Following DEXTR's training protocol \cite{maninis18} with 4 extreme points using our architecture, we observe that two unconstrained contour clicks allow for a similar accuracy (-1.4\%) while being more than twice as fast (Table \ref{tab:time_comparison}, \ref{tab:extreme_comparison}). We also observe experimentally a robustness to click location variation (Figure \ref{fig:noise}), which further validates unconstrained clicks as a flexible and cognitively easy option for interactive segmentation. Qualitatively, our approach seems robust to object shape variation, occlusion and dense scenes (Figure \ref{fig:predictions}).

Figure \ref{fig:curves} gives a comparison of the ability of our model to improve segmentation masks with an increase number of user clicks with other methods. Our pipeline is able to continuously improve IoU with an increasing number of clicks. We observe a larger gap against other methods on SBD \cite{bharath11} which may be due to bias as its train and test set are most resembling. Note that we do not include the GAIS method in the curve comparison as they use a synthetic dataset for training.

When compared against other contour based methods, UCP-Net enables for a significant drop in the number of needed user clicks to achieve a satisfactory segmentation on SBD, GrabCut and COCO MVal. 


\section{Conclusion}

With our generic contour based approach we have shown that unconstrained contour clicks enable for faster and more accurate segmentation thanks to fewer user clicks. We set a new state of the art of interactive segmentation on SBD, Berkeley, and COCO MVal. Our method is suitable for annotation purposes, enabling to label datasets requiring only a handful of user interaction. Moreover, it is also perfectly suitable for image editing applications as our iterative scheme makes it possible to reach a very high accuracy. In future work, investigating the contour clicks' embedding might prove relevant to best exploit this interaction as it was found for extreme clicks \cite{wang19}. Moreover, the usage of the previously predicted segmentation yields significant improvement in iterative positive and negative interaction approaches \cite{benard18,forte20} and may be equally applicable for unconstrained clicks. Given the nature of contour clicks, they could also be further exploited to simultaneously segment or correct objects which are close to one another or which overlap as they share common boundaries.




{\small
\bibliographystyle{ieeetr}
\bibliography{egbib}
}

\end{document}